# Comparative Performance of Machine Learning Architectures for Fault Detection and Diagnosis in Chemical Processes

R. AMMAR KHODJA [1, 2], A. VOISIN [2], V. COSTA [1], B. CELSE [1], F. CASTERAN [1], B. IUNG [2]

[1] IFPEN, Rond-point de l'échangeur de Solaize, France, *rayane.ammar-khodja@ifpen.fr*, *victor.costa@ifpen.fr*, *fanny.casteran@ifpen.fr*, *benoit.celse@ifpen.fr*

[2] *Université de Lorraine, UMR CNRS 7039, CRAN, F-54000, Nancy, France*, *alexandre.voisin@univ-lorraine.fr*, *benoit.iung@univ-lorraine.fr*

**Abstract**. Fault Detection and Diagnosis (FDD) is crucial for both maintenance and control in chemical industries, where early fault detection can prevent costly failures and optimize operations. This is particularly critical in pilot units like those at IFP Energies Nouvelles (IFPEN), which operate under short-term experimental conditions with frequently varying operational parameters.

This study conducts an extensive benchmarking analysis using the Tennessee Eastman Process (TEP), a widely used simulated chemical process dataset, to evaluate multiple approaches for fault detection and diagnosis, featuring numerous continuous operations with different sensors and faults. Several methods were implemented and compared including Multi Scale PCA (MS-PCA), AutoEncoder, Ensemble Learning, and LSTM models for fault detection, alongside Random Forest, XGBoost, and BLSTM (Bidirectional LSTM) for fault diagnosis.

Using the TEP dataset, our results demonstrate that Ensemble Learning achieves detection rates ranging from 80% to 100% across various fault scenarios for the fault detection task. For the fault diagnosis task, BLSTM achieved a diagnosis accuracy of 98.76%. The study reveals that ensemble-based approaches consistently outperform individual models in handling the complex, multivariate nature of chemical process data, due to its robustness in combining multiple perspectives, comprehensive data capture, and localized detection capabilities. Furthermore, the superior performance of the BLSTM is due to its ability to capture both past and future temporal dependencies in the sequential data, particularly important in chemical processes where fault patterns may manifest with complex temporal relationships.

This research contributes to the field of Fault Detection and Diagnosis by providing empirical evidence for the effectiveness of ensemble methods and Bidirectional LSTMs in order to address industrial FDD applications on chemical processes.

## 1. Introduction

Chemical processes present unique challenges for fault detection and diagnosis due to their inherently complex, nonlinear dynamics and strong interdependencies among process variables. In these environments, undetected faults can lead to product quality deterioration, equipment damage, environmental hazards, or catastrophic safety incidents with potential loss of life. Traditional FDD approaches in chemical processing have relied on first-principles models, statistical process control, and expert systems, which often require extensive process knowledge and struggle with the high dimensionality of modern chemical plants that may incorporate hundreds of sensors.

The evolution of artificial intelligence has introduced advanced computational methods that offer significant improvements over classical approaches. Machine learning and deep learning algorithms demonstrate remarkable capabilities in processing large volumes of data to identify latent patterns and adapt to novel anomaly types that conventional methods might overlook. This advancement is particularly valuable for chemical reactors, where early anomaly detection and accurate diagnosis are critical for forecasting undesired events and enhancing process efficiency.

While various machine learning approaches have been proposed for Fault Detection and Diagnosis (FDD) applications [Schmidl & Wenig], there remains a significant gap in comprehensive testing of ensemble models and deep learning architectures for this specific application [Zamanzadeh Darban et al, 2025]. These advanced methods have shown promise in other domains but require thorough evaluation in the context of chemical process monitoring, where data is often multivariate, non-linear, and exhibits complex temporal dependencies.

## 2. Material and Methods

### 2.1 *Tennessee Eastman Process Dataset*

The TEP dataset is derived from a real industrial process that was computationally modelled in 1993 by Downs and Vogel [Downs & Vogel, 1993]. It has become a widely used benchmark for studying industrial process control, particularly in FDD. The process involves 8 chemical components (4 reactants, 2 products, 1 by-product, and 1 inert component) and 5 main process units (reactor, condenser, separator, compressor, and stripper).

The dataset provides 53 different variables tracking system properties, including flow rates, pressures, temperatures, levels, mole fractions, and compressor power outputs. Process variables are sampled every 3 minutes for 25 hours in the training set and 48 hours in the testing set. The dataset contains both fault-free data and 20 different simulated process faults, making it ideal for benchmarking anomaly detection algorithms.

### 2.2 *Fault Detection Methods*

Anomaly detection in industrial processes involves identifying deviations from normal operating conditions by establishing baseline patterns of healthy system behaviour and detecting significant departures from these patterns. Most advanced methods employ a reconstruction-based approach [Zamanzadeh Darban et al, 2025], where models are trained on fault-free data to learn normal process behaviour, and anomalies are detected when the reconstruction error exceeds a predetermined threshold. In our study, we evaluated five distinct approaches:

- Statistical Thresholding (ST) establishes control limits at ±3 standard deviations from the mean for each sensor based on fault-free data, flagging values outside this range as potential faults.
- Multi-Scale PCA (MS-PCA) transforms high-dimensional process data into a lower-dimensional space to capture the principal patterns, detecting anomalies when the reconstruction error between original and PCA-reconstructed data exceeds a statistical threshold.
- AutoEncoder (AE) employs a neural network architecture to compress input data into a latent representation and then reconstruct it, identifying anomalies when the reconstruction error is abnormally high.
- Ensemble Learning (EL) constructs multiple linear regression models where each model predicts one specific feature based on all others, detecting faults when the collective prediction error exceeds expected values.
- Long Short-Term Memory (LSTM) networks analyse temporal sequences using a sliding-window approach to capture time-dependent patterns in the process data, recognizing anomalies when the predicted sequence deviates significantly from actual measurements.

### 2.3 *Fault Diagnosis Methods*

Fault diagnosis extends beyond detection to accurately classify and identify the specific type of fault occurring in the system, enabling appropriate corrective actions and maintenance procedures. Once an anomaly is detected, diagnosis techniques analyse the pattern of deviations across multiple sensors to determine the root cause from among possible fault categories. For this critical classification task, we evaluated three promising machine learning approaches:

- Random Forest Classifier constructs multiple decision trees during training and determines the fault class through majority voting, providing robust classification by aggregating predictions across a diverse ensemble of trees.
- XGBoost Classifier implements gradient boosting techniques that sequentially build decision trees to correct errors made by previous trees, creating a powerful ensemble that excels at capturing complex nonlinear relationships in process data.
- Bidirectional LSTM (BLSTM) processes time-series data in both forward and backward directions to capture temporal dependencies from both past and future contexts, making it particularly effective for fault patterns with complex sequential characteristics.

2.4 *Methodology*

All models were trained exclusively on fault-free data to learn normal process behavior patterns. For the reconstruction-based detection methods, data normalization was performed using z-score standardization to ensure equal contribution from each variable.

Hyperparameter optimization employed grid search with 5-fold cross-validation to identify optimal configurations. Neural networks benefited from an exponential decay learning rate schedule starting at $10^{-3}$ and decreasing by a factor of 0.9 every 1000 steps.

Detection thresholds were established at $\mu + 3\sigma$ of the reconstruction loss distribution from fault-free validation data, providing statistical significance at the 99.73% confidence level.

For classification models, training involved label encoding of the 20 fault types as categorical targets, an 80:20 train-validation split for hyperparameter tuning, and techniques to address the imbalanced nature of fault occurrences to prevent bias toward majority classes.

*2.4 Metrics used to compare methods*

Our comprehensive evaluation framework employed multiple complementary metrics to thoroughly assess both detection and diagnosis capabilities. For fault detection performance, we utilized Fault Detection Rate (FDR) to measure the proportion of correctly detected faults relative to all actual faults, providing insight into the models' sensitivity. Accuracy measured the overall correctness of predictions, counting both true positives and true negatives. To address potential imbalances in the dataset, we incorporated Matthews Correlation Coefficient (MCC), which provides a balanced assessment by considering all confusion matrix components even when classes are disproportionately represented.

Receiver Operating Characteristic (ROC) curves enabled visualization of the trade-off between true positive and false positive rates across varying detection thresholds, with Area Under the Curve (AUC) serving as a single comprehensive metric of detection quality.

For fault diagnosis evaluation, we employed confusion matrices to visualize classification performance across all fault types, highlighting both successful classifications and misclassifications between similar fault patterns. Classification accuracy provided an overall performance metric, while t-SNE dimensionality reduction enabled visualization of the models' internal representations, offering valuable insights into how effectively each approach separated different fault classes in the feature space.

## 3. Results and Analysis

3.1 *Fault Detection Results*

The performance of fault detection methods was evaluated using multiple metrics across 200-day simulations. Ensemble Learning demonstrated superior performance with accuracy rates ranging from 80% to 100% across most fault scenarios (Table 1). LSTM also showed strong performance, particularly after optimizing window size and stride parameters.

Analysis of the ROC curves (not showed here) revealed that choosing a threshold of $\mu + 3\sigma$ consistently resulted in the smallest False Positive Rate (FPR) and the highest True Positive Rate (TPR) for all five fault detection methods. This threshold effectively minimizes false alarms while maximizing the detection of actual faults.

The strengths of Ensemble Learning included robustness through combining multiple models, effective identification of deviations in specific sensors, and computational efficiency suitable for real-time monitoring. Additionally, it demonstrated superior performance in MCC scores, indicating balanced classification across various fault scenarios.

3.2 *Fault Diagnosis Results*

For fault classification over 400-day simulations, the evaluated models demonstrated increasing levels of accuracy, with Random Forest achieving 91%, XGBoost reaching 93%, and BLSTM delivering 99% accuracy.

The BLSTM model significantly outperformed both RF and XGBoost classifiers, demonstrating its superior capability to capture temporal dependencies and sequential patterns critical for accurate fault diagnosis. The t-SNE visualization of the BLSTM's intermediate layer outputs (Figure 1) showed clear clustering of different fault classes with minimal overlap, confirming the model's ability to learn highly discriminative features for fault classification.

| No. | Fault Description | ST | MS-PCA | AE | EL | LSTM |
|---|---|---|---|---|---|---|
| 1 | A/C feed ratio, B comp. const. (Stream 4) | 86.77 | 85.41 | 85.83 | **99.37** | 99.05 |
| 2 | B comp., A/C ratio const. (Stream 4) | 86.25 | 84.79 | 85.10 | **98.54** | 99.26 |
| 4 | Reactor cooling water inlet temp. | 87.08 | 85.41 | 85.94 | **99.69** | 98.52 |
| 5 | Condenser cooling water inlet temp. | 41.04 | 16.35 | 65.73 | **99.69** | 28.63 |
| 6 | A feed loss (Stream 1) | 87.08 | 85.62 | 86.04 | **99.69** | 98.63 |
| 7 | C header pressure loss (Stream 4) | 87.08 | 85.62 | 86.04 | **99.69** | 98.63 |
| 8 | A, B, and C comp. (Stream 4) | 85.41 | 80.10 | 84.16 | 96.25 | **97.89** |
| 10 | C feed temp. (Stream 4) | 49.27 | 19.06 | 19.48 | **66.14** | 20.74 |
| 11 | Reactor cooling water inlet temp. | 72.29 | 48.13 | 51.88 | 69.79 | **72.32** |
| 12 | Condenser cooling water inlet temp. | 86.45 | 82.91 | 84.38 | **98.13** | 98.11 |
| 13 | Reaction kinetics | 82.50 | 79.06 | 80.00 | 93.33 | **94.95** |
| 14 | Reactor cooling water valve | 87.08 | 79.06 | 85.94 | **99.69** | 98.63 |
| 16 | Unknown | 42.70 | 29.47 | 25.83 | **65.41** | 19.68 |
| 17 | Unknown | 77.71 | 75.72 | 76.66 | **89.16** | **89.16** |
| 18 | Unknown | 81.35 | 80.21 | 80.83 | 93.22 | **94.42** |
| 19 | Unknown | 42.29 | 8.85 | 10.41 | **69.37** | 22.84 |
| 20 | Unknown | 51.14 | 46.87 | 45.41 | **67.71** | 67.16 |

Table 1. Comparison of Fault Detection Methods using Accuracy (Percentage %).

Real-time fault classification using BLSTM further validated this performance, showing high-confidence predictions with distinct probability distributions for each fault type, reinforcing the model's reliability for industrial fault diagnosis applications.

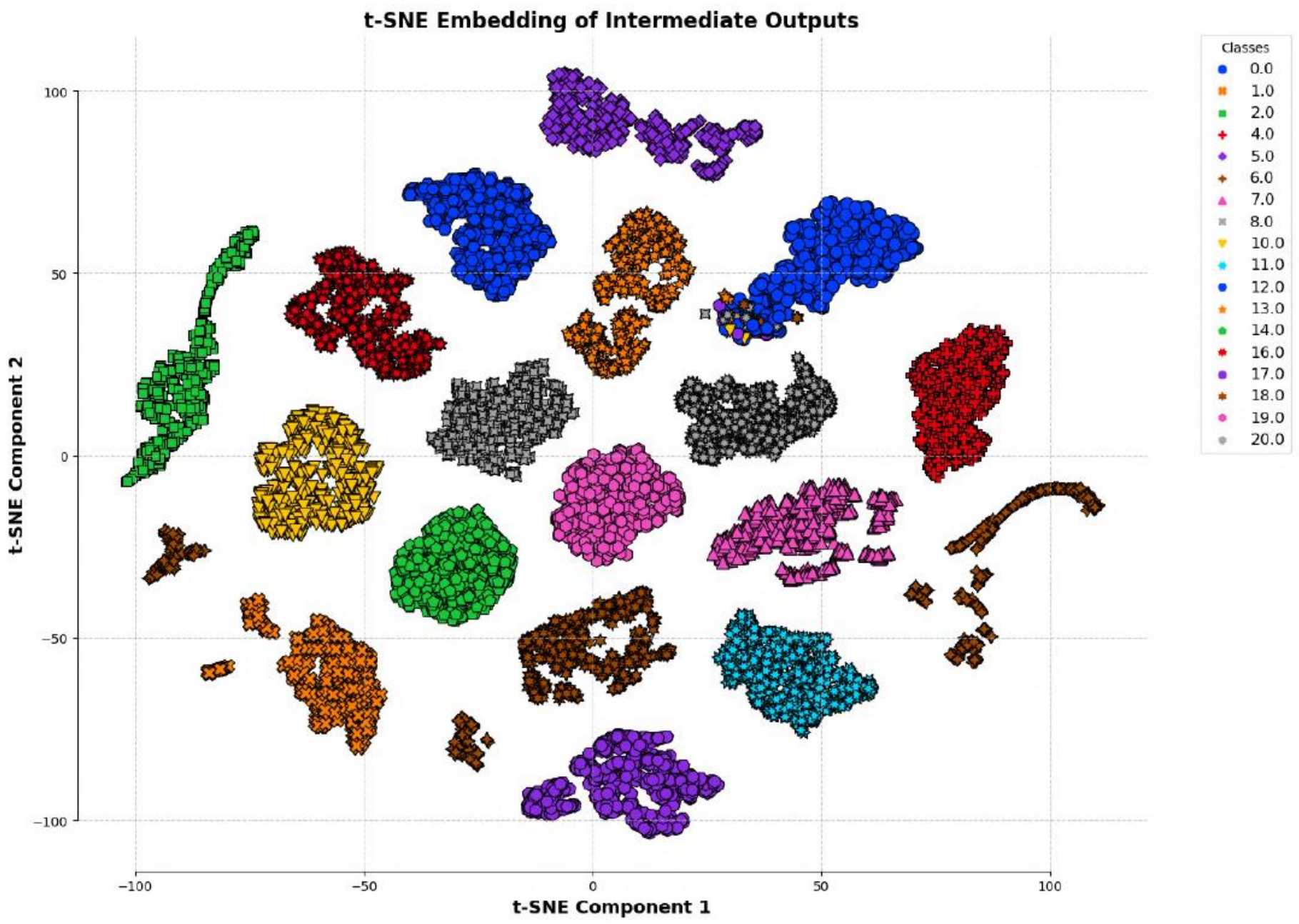

Figure 1. t-SNE plot for the classes of the trained data in the BLSTM for 400 days simulations.

To enhance fault interpretability and provide deeper diagnostic insights, we implemented a variable contribution analysis method for detected anomalies, as illustrated in Figure 2. This visualization approach not only identifies when a fault occurs but also determines which specific process variables contribute most significantly to the anomaly detection. The upper plot displays the reconstruction loss trajectory, clearly marking when it exceeds the detection threshold, while the lower plot quantifies the individual contribution of each sensor variable at a selected timestep after fault detection. This addition proves particularly valuable for maintenance personnel, as it pinpoints the most affected process components, effectively bridging the gap between detection and diagnosis. In the case shown, the reactor cooling water flow variable (xmvs_10) exhibits a substantially higher contribution compared to other variables, directly corresponding to the nature of Fault 4, which involves changes in reactor cooling water inlet temperature.

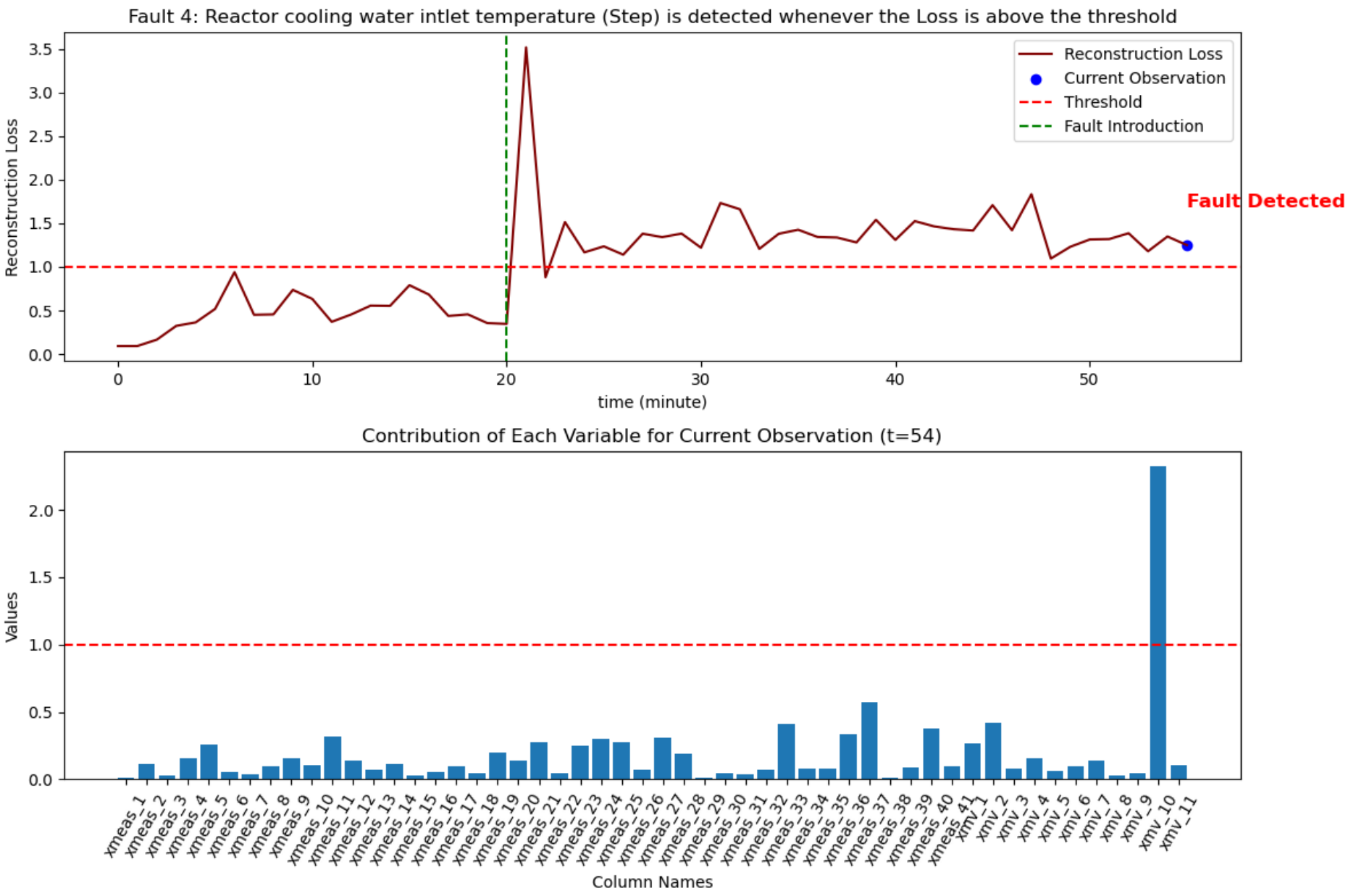

Figure 2. Anomaly detection in the Tennessee Eastman process: Analysis of reconstruction loss and identification of contributing variables: evolution of reconstruction loss over time (to identify when the fault was introduced) and analysis of individual variable contributions at time t=54 (162 minutes from the start of simulation) for anomaly diagnosis.

## 4. Conclusion

This study contributes to the field of fault detection and diagnosis by demonstrating the effectiveness of ensemble-based and deep learning approaches for industrial chemical processes. Our comprehensive evaluation framework established that ensemble methods for detection and bidirectional sequence models for diagnosis offer significant advantages in handling the complex, multivariate nature of process data. The complementary strengths of these approaches provide a foundation for robust fault management systems that can significantly enhance industrial safety and efficiency.

Future work will focus on developing predictive capabilities to forecast faults several days in advance and addressing the significant challenge of applying these methods to real pilot plants characterized by frequently changing operating conditions. Unlike production facilities, research pilot plants deliberately modify process parameters to explore new chemical pathways and optimizations, requiring FDD systems that can distinguish between intentional parameter changes and actual faults, while facing the additional challenge that real-life chemical pilot plants typically lack labeled data distinguishing between faulty and non-faulty operations or identifying specific fault types. This dynamic environment necessitates adaptive models capable of functioning effectively when baseline "normal" operation continuously evolves.